\documentclass[twoside,11pt]{article}

\usepackage[english]{babel}
\usepackage[utf8]{inputenc}
\usepackage{amsmath}
\usepackage{graphicx, subfigure}
\usepackage[colorinlistoftodos]{todonotes}
\usepackage{mlhc}
\usepackage{hyperref}
\usepackage{url}
\usepackage{amsfonts}
\usepackage{bbm}
\usepackage{appendix}
\usepackage{booktabs}
\ShortHeadings{Modeling Missing Data in Clinical Time Series with RNNs}{Lipton, Kale, and Wetzel}

\newcommand{\eat}[1]{}

\begin{document}

\title{Modeling Missing Data in Clinical Time Series with RNNs}

\author{\\
\name Zachary C. Lipton \email zlipton@cs.ucsd.edu \\
       \addr Department of Computer Science and Engineering \\
University of California, San Diego \\
La Jolla, CA 92093, USA
       \AND
       \name David C. Kale \email kale@isi.edu \\
       \addr USC Information Sciences Institute \\
       Marina del Rey, CA, USA
\AND
       \name Randall Wetzel \email rwetzel@chla.usc.edu \\
       \addr Laura P. and Leland K. Whittier Virtual Pediatric Intensive Care Unit \\
Children's Hospital LA \\
Los Angeles, CA 90089 \\
} 

\date{\today}

\maketitle


\begin{abstract}
We demonstrate a simple strategy 
to cope with missing data in sequential inputs, 
addressing the task of  multilabel classification 
of diagnoses given clinical time series. 
Collected from the pediatric intensive care unit (PICU) 
at Children's Hospital Los Angeles, 
our data consists of multivariate time series of observations. 
The measurements are irregularly spaced, 
leading to missingness patterns in temporally discretized sequences. 
While these artifacts are typically handled by imputation, 
we achieve superior predictive performance 
by treating the artifacts as features.
Unlike linear models, recurrent neural networks can realize this improvement using only simple binary indicators of missingness.
For linear models, 
we show an alternative strategy 
to capture this signal. 
Training models on missingness patterns only, 
we show that for some diseases, 
\emph{what tests are run} can be as predictive as the results themselves.
\end{abstract}

\section{Introduction}

For each admitted patient, hospital intensive care units record large amounts data
in electronic health records (EHRs).
Clinical staff routinely chart vital signs during hourly rounds and when patients are unstable.
EHRs record lab test results and medications as they are ordered or delivered by physicians and nurses.
As a result, EHRs contain rich
sequences of clinical observations depicting both patients' health and care received.
We would like to mine these time series
to build accurate predictive models
for diagnosis and other applications.
Recurrent neural networks (RNNs) are well-suited to learning sequential or temporal relationships 
from such time series.
RNNs offer unprecedented predictive power 
in myriad sequence learning domains,
including natural language processing, speech, 
video, and handwriting.
Recently, \citet{lipton2016learning} 
demonstrated the efficacy of RNNs 
for multilabel classification of diagnoses 
in clinical time series data.

However, medical time series data 
present modeling problems
not found in the clean academic datasets 
on which most RNN research focuses.
Clinical observations are recorded irregularly,
with measurement frequency varying between patients, 
across variables, and even over time.
In one common modeling strategy, 
we represent these observations as a sequence with discrete, fixed-width time steps. 
Problematically,
the resulting sequences often
contain missing values \citep{marlin:ihi2012}.
These values are typically not missing at random, 
but reflect decisions by caregivers.
Thus, the pattern of recorded measurements
contain potential information about the state of the patient.
However, most often, researchers fill missing values
using heuristic or unsupervised imputation \citep{lasko:plosone2013}, ignoring the potential predictive value of the missingness itself.

\begin{figure}[t]
  \centering
\includegraphics[clip=true,trim=0 150 0 90,width=0.85\textwidth]{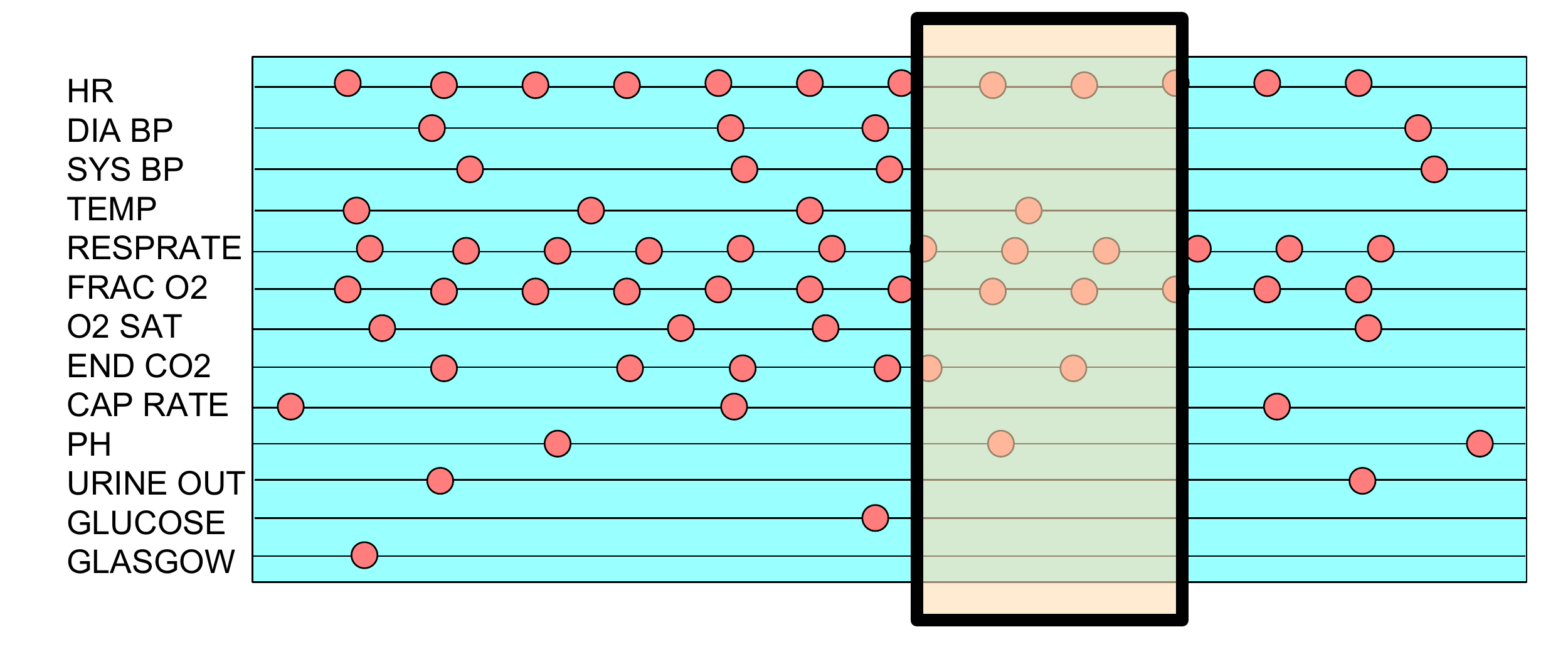}
  \caption{Missingness artifacts created by discretization}
\label{fig:missing-values}
\end{figure}

In this work we extend the methodology of \cite{lipton2016learning}
for RNN-based multilabel prediction of diagnoses.
We focus on data gathered from the Children's Hospital Los Angeles pediatric intensive care unit (PICU).
Unlike \cite{lipton2016learning}, 
who approach missing data via heuristic imputation,
we directly model missingness as a feature, achieving superior predictive performance. 
RNNs can realize this improvement using only simple binary indicators for missingness.
However, linear models are unable to use indicator features as effectively.
While RNNs can learn arbitrary functions, capturing the interactions between the missingness indicators the sequence of observation inputs, 
linear models can only learn substitution values. 
For linear models, 
we introduce an alternative strategy to capture this signal,
using a small number of simple hand-engineered features.

Our experiments demonstrate the benefit modeling missing data as a first-class feature.
Our methods improve the performance of RNNs, multilayer perceptrons (MLPs), and linear models.
Additionally we analyze the predictive value of missing data information by training models on the missingness indicators only.
We show that for several diseases, \emph{what tests are run} 
can be as predictive as 
the actual measurements.
While we focus on classifying diagnoses, our methods can be applied to any predictive modeling problem involving sequence data
and missing values, such as early prediction of sepsis \citep{henry2015targeted} or real-time risk modeling \citep{wiens2012patient}.

It is worth noting that
we may not want our predictive models 
to rely upon the patterns of treatment, 
as argued by \cite{caruana2015intelligible}.
Once deployed, our models may influence 
the treatment protocols, 
shifting the distribution of future data, 
and thus invalidating their predictions.
Nonetheless, doctors at present 
often utilize knowledge of past care, 
and treatment signal can leak into the actual measurements themselves in ways that sufficiently powerful models can exploit.
As a final contribution of this paper,
we present a critical discussion 
of these practical and philosophical issues.


\section{Data}
Our dataset consists of patient records
extracted from
the EHR system at CHLA \citep{marlin:ihi2012,che:kdd2015} 
as part of an IRB-approved study. 
In all, the dataset contains 
$10,401$ PICU episodes.
Each episode describes the stay of one patient in the PICU for a period of at least 12 hours.
In addition, each patient record 
contains a static set of diagnostic codes, 
annotated by physicians 
either during or after each PICU visit.

\subsection{Inputs}
In their rawest representation,
episodes consist of irregularly spaced measurements of 13 variables: 
diastolic and systolic blood pressure, peripheral capillary refill rate,
end-tidal CO$_2$ (ETCO$_2$), fraction of inspired O$_2$ (FIO$_2$),
total Glascow coma scale, blood glucose, heart rate, pH, 
respiratory rate, blood oxygen saturation,
body temperature, and urine output. 
To render our data suitable for learning with RNNs, 
we convert to discrete sequences of hourly time steps, 
where time step $t$ covers the interval 
between hours $t$ and $t+1$, 
closed on the left but open on the right. 
Because actual admission times are not recorded reliably, 
we use the time of the first recorded observation as 
time step $t=0$.
We combine multiple measurements of the same variable
within the same hour window by taking their mean.

Vital signs, such as heart rate, 
are typically measured about once per hour, 
while lab tests requiring a blood draw (e.g., glucose) 
are measured on the order of once per day 
(see appendix B
for measurement frequency statistics). 
In addition, the timing of and time between observations
varies across patients and over time. 
The resulting sequential representation
have many missing values, 
and some variables missing altogether.

Note that our methods can be sensitive to the duration of our discrete time step. 
For example, halving the duration
would double the length of the sequences, 
making learning by backpropagation through time more challenging \citep{bengio1994learning}. 
For our data, such cost would not be justified 
because the most frequently measured variables (vital signs) 
are only recorded about once per hour.
For higher frequency recordings of variables with faster dynamics, 
a shorter time step might be warranted.

To better condition our inputs, 
we scale each variable to the $[0,1]$ interval, 
using expert-defined ranges.
Additionally, we correct for differences in heart rate, respiratory rate, \citep{fleming2011lancet} and blood pressure \citeyearpar[NHBPEP Working Group][]{nhbp2004report} 
due to age and gender
using tables of normal values 
from large population studies.

\subsection{Diagnostic labels}
In this work, we formulate \textit{phenotyping} \citep{oellrich2015phenotyping}
as multilabel classification of sequences.
Our labels include $429$ distinct diagnosis codes from an in-house taxonomy at CHLA,
similar to 
ICD-9 codes \citep{world2004international}
commonly used in medical informatics research.
These labels include a wide range of acute conditions, such as acute respiratory distress, congestive heart failure, and sepsis. A full list is given in
appendix A.
We focus on the $128$ most frequent, each having at least 50 positive examples in our dataset.
Naturally, the diagnoses are not mutually exclusive. 
In our data set, the average patient 
is associated with $2.24$ diagnoses.
Additionally, the base rates of the diagnoses vary widely (see 
appendix A).
 


\section{Recurrent Neural Networks for Multilabel Classification}
While our focus in this paper is on missing data,
for completeness, we review the LSTM RNN
architecture for performing multilabel classification of diagnoses introduced by \cite{lipton2016learning}.
Formally,
given a series of observations $\boldsymbol{x}^{(1)},...,\boldsymbol{x}^{(T)}$,
we desire a classifier to generate hypotheses $\boldsymbol{\hat{y}}$
of the true labels $\boldsymbol{y}$,
where each input $\boldsymbol{x}^t \in \mathbbm{R}^D$ and the output $\boldsymbol{\hat{y}} \in [0,1]^K$.
Here, $D$ denotes the input dimension,
$K$ denotes the number of labels,
$t$ indexes sequence steps, 
and for any example, 
$T$ denotes the length of that sequence.

Our proposed RNN uses LSTM memory cells \citep{hochreiter1997long} 
with forget gates \citep{gers2000learning} 
but without peephole connections \citep{gers2003learning}. 
As output, we use a fully connected layer
followed by an element-wise logistic activation function $\sigma$.
We apply \emph{log loss} (binary cross-entropy) 
as the loss function at each output node.

The following equations give the update for a layer of memory cells $\boldsymbol{h}_l^{(t)}$, 
where $\boldsymbol{h}_{l-1}^{(t)}$ stands for the previous layer at the same sequence step (a previous LSTM layer or the input $\boldsymbol{x}^{(t)}$) 
and $\boldsymbol{h}^{(t-1)}_{l}$ stands for the same layer at the previous sequence step:
\begin{eqnarray*}
\boldsymbol{g}_l^{(t)} &=& \phi( W_l^{\mbox{gx}} \boldsymbol{h}^{(t)}_{l-1} +   W_l^{\mbox{gh}} \boldsymbol{h}^{(t-1)}_{l}  + \boldsymbol{b}_l^{\mbox{g}})\\
\boldsymbol{i}_l^{(t)} &=& \sigma( W_l^{\mbox{ix}} \boldsymbol{h}^{(t)}_{l-1} + W_l^{\mbox{ih}} \boldsymbol{h}^{(t-1)}_{l} + \boldsymbol{b}_l^{\mbox{i}})\\
\boldsymbol{f}_l^{(t)} &=& \sigma( W_l^{\mbox{fx}} \boldsymbol{h}^{(t)}_{l-1} + W_l^{\mbox{fh}} \boldsymbol{h}^{(t-1)}_{l} + \boldsymbol{b}_l^{\mbox{f}})\\
\boldsymbol{o}_l^{(t)} &=& \sigma( W_l^{\mbox{ox}} \boldsymbol{h}^{(t)}_{l-1} + W_l^{\mbox{oh}} \boldsymbol{h}^{(t-1)}_{l} + \boldsymbol{b}_l^{\mbox{o}})\\
\boldsymbol{s}_l^{(t)} &=& \boldsymbol{g}_l^{(t)} \odot \boldsymbol{i}_l^{(i)} + \boldsymbol{s}_l^{(t-1)} \odot \boldsymbol{f}_l^{(t)}\\
\boldsymbol{h}^{(t)}_{l} &=& \phi(\boldsymbol{s}_l^{(t)}) \odot \boldsymbol{o}_l^{(t)}
\end{eqnarray*}

In these equations, $\sigma$ stands for an element-wise application of the \textit{logistic} function, 
$\phi$ stands for an element-wise application 
of the $tanh$ function, 
and $\odot$ is the Hadamard (element-wise) product. 
The input, output, and forget gates 
are denoted by $\boldsymbol{i}$, 
$\boldsymbol{o}$, and $\boldsymbol{f}$ respectively, while $\boldsymbol{g}$ is the input node 
and has a \textit{tanh} activation.

The loss at a single sequence step 
is the average \emph{log loss} calculated across all labels:
$$
\mbox{loss}(\boldsymbol{\hat{y}}, \boldsymbol{y}) = \frac{1}{K} \sum_{l=1}^{l=K} - (y_l \cdot \mbox{log}(\hat{y}_l) + (1 - y_l) \cdot \mbox{log}(1 - \hat{y}_l) ) .
$$
%
To overcome the difficulty 
of learning to pass information 
across long sequences,
we use the \textit{target replication}
strategy proposed by \citet{lipton2016learning}, in which
we replicate the static targets 
at each sequence step 
providing a local error signal.
This technique is also motivated
by our problem: we desire to make accurate predictions  
even if the sequence were truncated (as in
early-warning systems).
To calculate loss, we take a convex combination
of the final step loss and the average 
of the losses over predictions $\boldsymbol{\hat{y}}^{(t)}$ at all steps $t$:
$$\alpha \cdot \frac{1}{T} \sum_{t=1}^{T} \mbox{loss}(\boldsymbol{\hat{y}}^{(t)}, \boldsymbol{y}^{(t)})
+ (1-\alpha) \cdot \mbox{loss}(\boldsymbol{\hat{y}}^{(T)}, \boldsymbol{y}^{(T)})$$
where $\alpha \in [0,1]$ is a hyper-parameter
determining the relative importance 
of performance on the intermediary vs. final targets.
At inference time, 
we consider only the output at the final step.



\section{Missing Data}
\label{sec:missing}
In this section, we explain our procedures for imputation, missing data indicator sequences, engineering features of missing data patterns. 

\subsection{Imputation}
To address the missing data problem, 
we consider two different imputation strategies 
(forward-filling and zero imputation), 
as well as direct modeling via indicator variables.
Because imputation and direct modeling are not mutually exclusive,
we also evaluate them in combination.
Suppose that $x_i^{(t)}$
is ``missing.''
In our \textit{zero-imputation} strategy, we simply set $x_i^{(t)} := 0$ whenever it is missing.
In our \textit{forward-filling} strategy, we impute $x_i^{(t)}$ as follows:
\begin{itemize}
\item If there is at least one previously recorded measurement of variable $i$ at a time $t'<t$, 
we perform forward-filling by setting $x_i^{(t)} := x_i^{(t')}$.
\item If there is no previous recorded measurement (or if the variable is missing entirely), 
then we impute the median estimated over all measurements in the training data.
\end{itemize}
This strategy is motivated by the intuition 
that clinical staff record measurements 
at intervals proportional to rate 
at which they are believed or observed to change. 
Heart rate, which can change rapidly, 
is monitored much more frequently than blood pH.
Thus it seems reasonable 
to assume that a value has changed little 
since the last time it was measured.


\begin{figure}[ht]
  \begin{center}
	\includegraphics[clip=true,trim=0 50 0 50,width=0.8\textwidth]{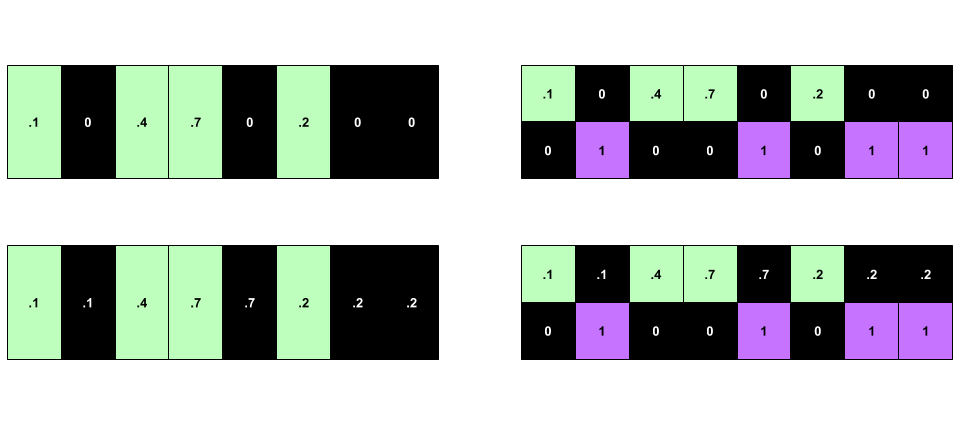}
  \end{center}
  \caption{(top left) no imputation or indicators, (bottom left) imputation absent indicators, (top right) indicators but no imputation, (bottom right) indicators and imputation. Time flows from left to right.}
\label{fig:indicators_imputation}
\vspace{-30px}
\end{figure} 

\subsection{Learning with Missing Data Indicators}
\label{sec:indicators}
Our \textit{indicator variable} approach to missing data consists of augmenting our inputs with binary variables $m^{(t)}_i$ 
for every $x_i^{(t)}$, where $m^{(t)}_i := 1$ if $x_i^{(t)}$ is imputed and 0 otherwise.
Through their hidden state computations, 
RNNs can use these indicators 
to learn arbitrary functions of the past observations and missingness patterns.
However, given the same data,
linear models can only learn hard substitution rules.  
To see why, consider a linear model
that outputs prediction $f(z)$, 
where $z = \sum_i w_i \cdot x_i$.
With indicator variables, 
we might say that
$z = \sum_i w_i \cdot x_i + \sum_i \theta_i \cdot m_i$
where
$\theta_i$ are the weights for each $m_i$.
If $x_i$ is set to $0$ and $m_i$ to $1$,
whenever the feature $x_i$ is missing,
then the impact on the output $\theta_i \cdot m_i = \theta_i$
is exactly equal to the contribution $w_i \cdot x^*_i$
for some $x^*_i = \theta_i / w_i$.
In other words, the linear model can only use the indicator
in a way that depends neither on the previously observed values ($x_i^{1}...x_i^{t-1}$), 
nor any other evidence in the inputs.
\begin{figure}[ht]
  \centering
\includegraphics[clip=true,trim=0 50 0 45,width=0.85\textwidth]{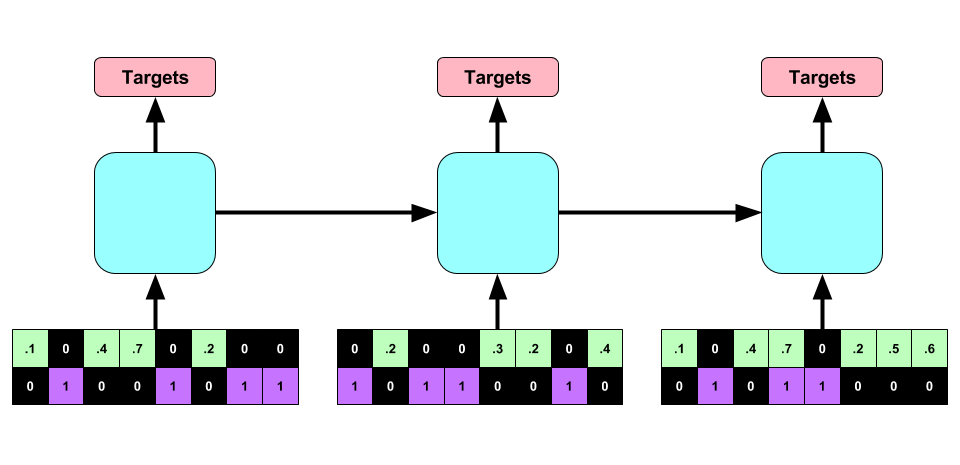}
  \caption{Depiction of RNN zero-filled inputs and missing data indicators.}
\label{fig:indicators_rnn}
\end{figure}

Note that for a linear model,
the impact of a missing data indicator on predictions
must be monotonic.
In contrast, the RNN might infer that for one patient heart rate is missing because they went for a walk,
while for another it might signify an emergency.
Also note that even without indicators, 
the RNN might learn to recognize \emph{filled-in}
vs \emph{real} values.
For example, with forward-filling, the RNN
could learn to recognize exact repeats.
For zero-filling, the RNN could recognize that values set to exactly $0$ were likely missing measurements.

\subsection{Hand-engineered missing data features}
To overcome the limits of the linear model,
we also designed features from the indicator sequences.  
As much as possible, we limited ourselves to features that are simple to calculate,
intuitive, and task-agnostic.
The first is a binary indicator for whether a variable was measured \textit{at all}.
Additionally, we compute 
the mean and standard deviation of the indicator sequence.
The mean captures the frequency with which each variable is measured which carries information about the severity of a patient's condition. 
The standard deviation, on the other hand,
computes a non-monotonic function of frequency that is maximized when a variable is missing exactly 50\% of the time. 
We also compute the frequency with which a variable switches from measured to missing or vice versa across adjacent sequence steps.
Finally, we add features that capture the relative timing of the first and last measurements of a variable, computed as the number of hours until the measurement divided by the length of the full sequence.

\section{Experiments}
We now present the training details and empirical findings of our experiments.
Our LSTM RNNs each have $2$ hidden layers of $128$ LSTM cells each,
non-recurrent dropout of $0.5$, and $\ell_2^2$ weight decay of $10^{-6}$.
We train on $80\%$ of data, setting aside $10\%$ each for validation and testing. 
We train each RNN for $100$ epochs,
retaining the parameters corresponding to the epoch with the lowest validation loss.

We compare the performance of RNNs against logistic regression and multilayer perceptrons (MLPs). 
We apply $\ell_2$ regularization to the logistic regression model. 
The MLP has $3$ hidden layers with $500$ nodes each, rectified linear unit activations, and dropout (with probability of 0.5), choosing the number of layers and nodes by validation performance. 
We train the MLP
using
stochastic gradient descent with momentum.

We evaluate each baseline with two sets of features: raw and hand-engineered.
Note that our baselines cannot be applied directly to variable-length inputs.
For the raw features, we concatenate three 12-hour subsequences, one each from the beginning, middle, and end of the time series.
For shorter time series, these intervals may overlap. Thus raw representations contain $2 \times 3 \times 12 \times 13 = 936$ features.
We train each baseline on five different combinations of raw inputs:
(1) measurements with zero-filling,
(2) measurements with forward-filling,
(2) measurements with zero-filling + missing data indicators,
(4) forward-filling + missing data indicators, and
(5) missing data indicators only.

Our hand-engineered features capture central tendencies, variability, extremes, and trends. These include the first and last measurements and their difference, maximum and minimum values, mean and standard deviation, median and 25th and 75th percentiles, and the slope and intercept of least squares line fit. We also computed the 8 missing data features described in \autoref{sec:missing}.
We improve upon the baselines in \citet{lipton2016learning} by computing the hand-engineered features over different windows of time, giving them access to greater temporal information and enabling them to better model patterns of missingness. 
We extract hand-engineered features from the entire time series and from three possibly overlapping intervals: the first and last 12 hours and the interval between (for shorter sequences, we instead use the middle 12 hours).
This yields a total of $4 \times 12 \times 13 = 624$ and $4 \times 8 \times 13 = 416$ hand-engineered measurement and missing data features, respectively. 
We train baseline models on three different combinations of hand-engineered features: (1) measurement-only, (2) indicator-only, and (3) measurement and indicator.

We evaluate all models on the same training, validation, and test splits.
Our evaluation metrics include area under the ROC curve (AUC) and F1 score (with threshold chosen based on validation performance).
We report both micro-averaged (calculated across all predictions) and macro-averaged (calculated separately on each label, then averaged) measures to mitigate the weaknesses in each \citep{lipton2014optimal}.
Finally we also report precision at $10$, 
whose maximum is $0.2238$ because we have on average $2.238$ diagnoses per patient.
This metric seems appropriate because we could imagine this technology would be integrated into a diagnostic assistant.
In that case, its role might be to suggest the most likely diagnoses among which a professional doctor would choose.
Precision at 10 evaluates the quality of the top 10 suggestions.

\begin{table*}[hbt!]
\centering
\footnotesize
\begin{tabular}{llllll}
\multicolumn{6}{c}{\textbf{ 
Classification performance for 128 ICU phenotypes}} \\
\toprule

\textbf{\textbf{Model}} & \textbf{Micro AUC} & \textbf{Macro AUC} & \textbf{Micro F1} & \textbf{Macro F1} & \textbf{P@10} \\
\midrule

\textbf{Base Rate} & 0.7128 & 0.5 & 0.1346 & 0.0343 & 0.0788 \\
\textbf{Best Possible} & 1.0 & 1.0 & 1.0 & 1.0 & 0.2281 \\

\midrule
\multicolumn{6}{c}{\textbf{Logistic Regression}} \\
\midrule
\textbf{Log Reg - Zeros } & 0.8108 & 0.7244 & 0.2149 & 0.0999 & 0.1014 \\
\textbf{Log Reg - Impute} & 0.8201 & 0.7455 & 0.2404 & 0.1189 & 0.1038 \\
\textbf{Log Reg - Zeros \& Indicators} & 0.8143 & 0.7269 & 0.2239 & 0.1082 & 0.1017 \\
\textbf{Log Reg - Impute \& Indicators} & 0.8242 & 0.7442 & 0.2467 & 0.1234 & 0.1045 \\
\textbf{Log Reg - Indicators Only} & 0.7929 & 0.6924 & 0.1952 & 0.0889 & 0.0939 \\


\midrule
\multicolumn{6}{c}{\textbf{Multilayer Perceptron}} \\
\midrule
\textbf{MLP - Zeros} & 0.8263 & 0.7502 & 0.2344 & 0.1072 & 0.1048 \\
\textbf{MLP - Impute} & 0.8376 & 0.7708 & 0.2557 & 0.1245 & 0.1031 \\
\textbf{MLP - Zeros \& Indicators} & 0.8381 & 0.7705 & 0.2530 & 0.1224 & 0.1067 \\

\textbf{MLP - Impute \& Indicators} & 0.8419 & 0.7805 & 0.2637 & 0.1296 & 0.1082 \\
\textbf{MLP - Indicators Only} & 0.8112 & 0.7321 & 0.1962 & 0.0949 & 0.0947 \\


\midrule
\multicolumn{6}{c}{\textbf{LSTMs}} \\
\midrule
\textbf{LSTM - Zeros} & 0.8662 & 0.8133 & 0.2909 & 0.1557 & 0.1176 \\
\textbf{LSTM - Impute} & 0.8600 & 0.8062 & 0.2967 & 0.1569 & 0.1159 \\
\textbf{LSTM - Zeros \& Indicators} & \textbf{0.8730} & \textbf{0.8250} & \textbf{0.3041} & \textbf{0.1656} & \textbf{0.1215} \\
\textbf{LSTM - Impute \& Indicators} & 0.8689 & 0.8206 & 0.3027 & 0.1609 & 0.1196 \\
\textbf{LSTM - Indicators Only} & 0.8409 & 0.7834 & 0.2403 & 0.1291 & 0.1074 \\


\midrule
\multicolumn{6}{c}{\textbf{Models using Hand-Engineered Features}} \\
\midrule
\textbf{Log Reg HE} & 0.8396 & 0.7714 & 0.2708 & 0.1327 & 0.1118 \\
\textbf{Log Reg HE Indicators} & 0.8472 & 0.7752 & 0.2841 & 0.1376 & 0.1165 \\
\textbf{Log Reg HE Indicators Only} & 0.8187 & 0.7322 & 0.2287 & 0.1001 & 0.1020 \\


\textbf{MLP HE} & 0.8599 & 0.8052 & 0.2953 & 0.1556 & 0.1168 \\
\textbf{MLP HE Indicators} & 0.8669 & 0.8160 & 0.2954 & 0.1610 & 0.1202 \\
\textbf{MLP HE Indicators Only} & 0.8371 & 0.7682 & 0.2351 & 0.1179 & 0.1028 \\

\bottomrule
\end{tabular} \hspace{-0.2in}
\caption{Performance on aggregate metrics for logistic regression (Log Reg), MLP, and LSTM classifiers with and without imputation and missing data indicators. }
\label{tab:experimental-results}
\vspace{-20px}
\end{table*}

\subsection{Results}
The best overall model by all metrics 
(micro AUC of $0.8730$) 
is an LSTM with zero-imputation and missing data indicators. 
It outperforms both the strongest MLP baseline
and LSTMs absent missing data indicators.
For the LSTMs using either imputation strategy, adding the missing data indicators improves performance in all metrics.
While all models improve with access to missing data indicators, this information confers less benefit to the raw input linear baselines, consistent with theory discussed in \autoref{sec:indicators}.

The results achieved by logistic regression with hand-engineered features indicates that our simple hand-engineered missing data features do a reasonably good job of capturing important information that neural networks are able to mine automatically.
We also find that LSTMs (with or without indicators) 
appear to perform better with zero-filling 
than with with imputed values. 
Interestingly, this is not true for either baseline. 
It suggests that the LSTM may be learning to recognize missing values implicitly by recognizing a tight range about the value zero and inferring that this is a missing value. 
If this is true, perhaps imputation interferes 
with the LSTM's ability to implicitly recognize missing values.
Overall, the ability to implicitly infer missingness 
may have broader implications.
It suggests that we might never completely hide this information from a sufficiently powerful model.



\section{Related Work}
This work builds upon research relating to missing values 
and machine learning for medical informatics.
The basic RNN methodology for phenotyping
derives from \citet{lipton2016learning}, 
addressing a dataset and problem described by \cite{che:kdd2015}.
The methods rely upon LSTM RNNs \citep{hochreiter1997long,gers2000learning}
trained by backpropagation through time \citep{Hinton06, werbos1988generalization}.
A comprehensive perspective on the history 
and modern applications of RNNs
is provided by \citet{lipton2015critical}, while \citet{lipton2016learning} list many of the previous works that have applied neural networks to digital health data.


While a long and rich literature addresses pattern recognition with missing data \citep{cohen1975applied,allison2001missing},
most of this literature addresses fixed-length feature vectors
\citep{garcia2010pattern,pigott2001review}.
Indicator variables for missing data 
were first proposed by \cite{cohen1975applied},
but we could not find papers that combine missing data indicators
with RNNs.
Only a handful of papers address missing data in the context of RNNS. 
\citet{bengio1996recurrent} demonstrate a scheme by which the RNN learns to fill in the missing values such that the filled-in values minimize output error. 
In 2001, \cite{parveen2001speech} built upon this method to improve automatic speech recognition.
 \citet{barker2001linking} suggests 
 using a mask of indicators in a scheme 
 for weighting the contribution of reliable vs corrupted data in the final prediction.
\citet{tresp:nips1997} address missing values by combining an RNN with a linear state space model to handle uncertainty.
This paper may be one of the first to 
engineer explicit features of missingness patterns in order to improve discriminative performance.
Also, to our knowledge, we are the first to harness patterns of missing data to improve the classification of critical care phenotypes.
\vspace{-10px}





\section{Discussion} 
\label{sec:discussion}
Data processing and discriminative learning 
have often been regarded as separate disciplines.
Through this separation of concerns, 
the complementarity of missing data indicators and training RNNs for classification has been overlooked. 
This paper proposes that patterns of missing values are an underutilized source of predictive power
and that RNNs, unlike linear models, 
can effectively mine this signal 
from sequences of indicator values.
Our hypotheses are confirmed by empirical evidence.
Additionally, we introduce and confirm the utility
of a simple set of features, engineered from the sequence of missingness indicators, 
that can improve performance of linear models.
These techniques are simple to implement and broadly applicable
and seem likely to confer similar benefits
on other sequential prediction tasks,
when data is missing not at random.
One example might include financial data,
where failures to report accounting details
could suggest internal problems at a company.


\subsection{The Perils and Inevitability of Modeling Treatment Patterns}
For medical applications, 
the predictive power of missing data 
raises important philosophical concerns.
We train models with supervised learning,
and verify their utility by assessing the accuracy of their classifications on hold-out test data.
However, in practice,
we hope to make treatment decisions
based on these predictions,
exposing a fundamental incongruity between the problem on which our models are trained and those for which they are ultimately deployed.
As articulated in 
\citet{lipton2016mythos}, 
these supervised models, trained offline, 
cannot account for changes that their deployment might confer upon the real world, 
possibly invalidating their predictions.
\cite{caruana2015intelligible} present a compelling case in which a pneumonia risk model predicted a lower risk of death 
for patients who also have asthma. 
The better outcomes of the asthma patients, 
as it turns out, owed to the more aggressive treatment they received. The model, if deployed,
might be used to choose less aggressive treatment
for the patients with both pneumonia and asthma,
clearly a sub-optimal course of action.

On the other hand,  to some degree, 
learning from treatment signal 
may be inevitable. 
Any imputation might leak some information 
about which values are likely imputed and which are not. Thus any sufficiently powerful supervised model might catch on to some amount of missingness signal, as was the case in our experiments with the LSTM using zero-filled missing values.
Even physiologic measurements 
contain information owing to patterns of treatment,
possibly reflecting
the medications patients receive and the procedures they undergo.

Sometimes the patterns of treatments 
may be a reasonable and valuable source of information.
Doctors already rely on this kind of signal habitually:
they read through charts, noting which other doctors have seen a patient, 
inferring what their opinions might have been from which tests they ordered.
While, in some circumstances, it may be problematic 
for learning models to rely on this signal,
removing it entirely may be both difficult and undesirable.

\subsection{Complex Models or Complex Features?}
Our work also shows that using only simple features, 
RNNs can achieve state of the art performance classifying clinical time series.
The RNNs far outperform linear models.
Still, in our experience, 
there is a strong bias among practitioners 
toward more familiar models even when they require substantial feature engineering.

In our experiments, we undertook extensive efforts 
to engineer features to boost the performance 
of both linear models and MLPs. 
Ultimately, while RNNs performed best on raw data, 
we could approach its performance with an MLP
and significantly improve the linear model
by using hand-engineered features and windowing. 
A question then emerges: 
how should we evaluate the trade-off 
between more complex models and more complex features? 
To the extent that linear models 
are believed to be more interpretable than neural networks, 
most popular notions of interpretability 
hinge upon the intelligibility of the features \citep{lipton2016mythos}.
When performance of the linear model  
comes at the price of this intelligibility,
we might ask if this trade-off undermines the linear model's chief advantage.
Additionally, such a model, while still inferior to the RNN, relies on application-specific features 
less likely to be useful on other datasets and tasks.
In contrast, RNNs seem better equipped 
to generalize to different tasks. 
While the model may be complex, 
the inputs remain intelligible, 
opening the possibility to various post-hoc interpretations \citep{lipton2016mythos}.

\subsection{Future Work}
We see several promising next steps following this work. 
First, we would like to validate this methodology on tasks with more immediate clinical impact, such as predicting sepsis, mortality, or length of stay.
Second, we'd like to extend this work towards predicting clinical decisions. 
Called policy imitation in the reinforcement literature, 
such work could pave the way to providing real-time decision support. 
Finally, we see machine learning as cooperating with a human decision-maker. 
Thus a machine learning model needn't always make a prediction/classification; it could also abstain. 
We hope to make use of the latest advances in mining uncertainty information from neural networks to make confidence-rated predictions.




\section{Acknowledgments}
Zachary C. Lipton was supported by the Division of Biomedical Informatics at the University of California, San Diego, via training grant (T15LM011271) from the NIH/NLM.
David Kale was supported by the Alfred E. Mann Innovation in Engineering Doctoral Fellowship.
The VPICU was supported by grants from the Laura P. and Leland K. Whittier Foundation.
We acknowledge NVIDIA Corporation for Tesla K40 GPU hardware donation and Professors Charles Elkan and Greg Ver Steeg for their support and advice. 

\bibliography{main}

\clearpage
\appendix
\section{Per Diagnosis Classification Performance}
\label{sec:bigtable}
In this appendix, we provide per-diagnosis AUC and F1 scores for three representative LSTM models trained with imputed measurements, with imputation plus missing indicators, and with indicators only. By comparing performance on individual diagnoses, we can gain some insight into the relationship between missing values and different conditions. Rows are sorted in descending order based on the F1 score of the imputation plus indicators model. It is worth noting that F1 scores are sensitive to threshold, which we chose in order to optimize per-disease validation F1, sometimes based on a very small number of positive cases. Thus, there are cases where one model will have superior AUC but worse F1.\\
\vspace{5pt}

\begin{table*}[ht]
\centering
\scriptsize
\begin{tabular}{lccccccc}
\multicolumn{8}{c}{\textbf{Classifier Performance on Each Diagnostic Code, Sorted by F1}} \\
\toprule
& & \multicolumn{2}{c}{\textbf{Msmt.}} & \multicolumn{2}{c}{\textbf{Msmt. + indic.}} & \multicolumn{2}{c}{\textbf{Indic.}} \\
\textbf{Condition} & Base rate & AUC & F1 & AUC & \textcolor{red}{\textbf{F1}} & AUC & F1 \\
\midrule
Diabetes mellitus with ketoacidosis & 0.0125 & 1.0000 & 0.8889 & 0.9999 & \textcolor{red}{\textbf{0.9333}} & 0.9906 & 0.7059 \\
Asthma with status asthmaticus & 0.0202 & 0.9384 & \textcolor{red}{\textbf{0.6800}} & 0.8907 & 0.6383 & 0.8652 & 0.5417 \\
Scoliosis (idiopathic) & 0.1419 & 0.9143 & \textcolor{red}{\textbf{0.6566}} & 0.8970 & 0.6174 & 0.8435 & 0.5235 \\
Tumor, cerebral & 0.0917 & 0.8827 & \textcolor{red}{\textbf{0.5636}} & 0.8799 & 0.5560 & 0.8312 & 0.4627 \\
Renal transplant, status post & 0.0122 & 0.9667 & 0.2963 & 0.9544 & 0.4762 & 0.9490 & \textcolor{red}{\textbf{0.5600}} \\
Liver transplant, status post & 0.0106 & 0.7534 & 0.3158 & 0.8283 & \textcolor{red}{\textbf{0.4762}} & 0.8271 & 0.2581 \\
Acute respiratory distress syndrome & 0.0193 & 0.9696 & 0.3590 & 0.9705 & \textcolor{red}{\textbf{0.4557}} & 0.9361 & 0.3333 \\
Developmental delay & 0.0876 & 0.8108 & \textcolor{red}{\textbf{0.4382}} & 0.8382 & 0.4331 & 0.6912 & 0.2366 \\
Diabetes insipidus & 0.0127 & 0.9220 & 0.2727 & 0.9486 & \textcolor{red}{\textbf{0.4286}} & 0.9266 & 0.4000 \\
End stage renal disease (on dialysis) & 0.0241 & 0.8548 & 0.2778 & 0.8800 & 0.4186 & 0.9043 & \textcolor{red}{\textbf{0.4255}} \\
Seizure disorder & 0.0816 & 0.7610 & 0.3694 & 0.7937 & \textcolor{red}{\textbf{0.4059}} & 0.6431 & 0.1957 \\
Acute respiratory failure & 0.0981 & 0.8414 & 0.4128 & 0.8391 & 0.3835 & 0.8358 & \textcolor{red}{\textbf{0.4542}} \\
Cystic fibrosis & 0.0076 & 0.8628 & 0.2353 & 0.8740 & \textcolor{red}{\textbf{0.3810}} & 0.8189 & 0.0000 \\
Septic shock & 0.0316 & 0.8296 & 0.3363 & 0.8911 & \textcolor{red}{\textbf{0.3750}} & 0.8506 & 0.1429 \\
Respiratory distress, other & 0.0716 & 0.8411 & \textcolor{red}{\textbf{0.3873}} & 0.8502 & 0.3719 & 0.7857 & 0.2143 \\
Intracranial injury, closed & 0.0525 & 0.8886 & 0.2817 & 0.9002 & \textcolor{red}{\textbf{0.3711}} & 0.8442 & 0.3208 \\
Arteriovenous malformation & 0.0223 & 0.8620 & 0.3590 & 0.8716 & \textcolor{red}{\textbf{0.3704}} & 0.8494 & 0.2857 \\
Seizures, status epilepticus & 0.0348 & 0.8381 & \textcolor{red}{\textbf{0.4158}} & 0.8505 & 0.3704 & 0.8440 & 0.3226 \\
Pneumonia due to adenovirus & 0.0123 & 0.8604 & 0.1250 & 0.9065 & \textcolor{red}{\textbf{0.3077}} & 0.8792 & 0.1818 \\
Leukemia (acute, without remission) & 0.0287 & 0.8585 & 0.2783 & 0.8845 & \textcolor{red}{\textbf{0.3059}} & 0.8551 & 0.2703 \\
Dissem. intravascular coagulopathy & 0.0099 & 0.9556 & \textcolor{red}{\textbf{0.5000}} & 0.9532 & 0.2857 & 0.9555 & 0.2500 \\
Septicemia, other & 0.0240 & 0.8586 & 0.2400 & 0.8870 & \textcolor{red}{\textbf{0.2812}} & 0.7593 & 0.0000 \\
Bronchiolitis & 0.0162 & 0.9513 & 0.2667 & 0.9395 & \textcolor{red}{\textbf{0.2703}} & 0.8826 & 0.1778 \\
Congestive heart failure & 0.0133 & 0.8748 & 0.1429 & 0.8756 & \textcolor{red}{\textbf{0.2703}} & 0.8326 & 0.1364 \\
Upper airway obstruc. (UAO), other & 0.0378 & 0.8206 & \textcolor{red}{\textbf{0.2564}} & 0.8573 & 0.2542 & 0.8350 & 0.1964 \\
Diabetes mellitus type I, stable & 0.0064 & 0.7105 & 0.0000 & 0.9625 & 0.2500 & 0.9356 & \textcolor{red}{\textbf{0.3333}} \\
Cerebral palsy (infantile) & 0.0262 & 0.8230 & \textcolor{red}{\textbf{0.2609}} & 0.8359 & 0.2500 & 0.6773 & 0.0980 \\
Coagulopathy & 0.0131 & 0.7501 & 0.1111 & 0.8098 & \textcolor{red}{\textbf{0.2449}} & 0.8548 & 0.1667 \\
UAO, ENT surgery, post-status & 0.0302 & 0.9059 & \textcolor{red}{\textbf{0.4058}} & 0.8733 & 0.2400 & 0.8364 & 0.1975 \\
Hypertension, systemic & 0.0169 & 0.8740 & 0.2105 & 0.8831 & 0.2388 & 0.8216 & \textcolor{red}{\textbf{0.2857}} \\
Acute renal failure, unspecified & 0.0191 & 0.9242 & 0.2381 & 0.9510 & 0.2381 & 0.9507 & \textcolor{red}{\textbf{0.3291}} \\
Trauma, vehicular & 0.0308 & 0.8673 & 0.2105 & 0.8649 & \textcolor{red}{\textbf{0.2381}} & 0.8022 & 0.1395 \\
Hepatic fail. (acute necrosis of liver) & 0.0176 & 0.8489 & 0.2222 & 0.9239 & \textcolor{red}{\textbf{0.2308}} & 0.8598 & 0.1935 \\
Craniosynostosis (anomalies of skull) & 0.0064 & 0.7824 & 0.0000 & 0.9267 & \textcolor{red}{\textbf{0.2286}} & 0.8443 & 0.0315 \\
Prematurity ($<$37 weeks gestation) & 0.0321 & 0.7520 & 0.1548 & 0.7542 & \textcolor{red}{\textbf{0.2245}} & 0.7042 & 0.1266 \\
Hydrocephalus, other (congenital) & 0.0381 & 0.7118 & 0.2099 & 0.7500 & \textcolor{red}{\textbf{0.2241}} & 0.7065 & 0.1961 \\
Pneumothorax & 0.0134 & 0.8220 & 0.1176 & 0.7957 & 0.2188 & 0.7552 & \textcolor{red}{\textbf{0.3243}} \\
Congenital muscular dystrophy & 0.0121 & 0.8427 & \textcolor{red}{\textbf{0.2500}} & 0.8491 & 0.2143 & 0.7460 & 0.0800 \\
Cardiomyopathy (primary) & 0.0191 & 0.7508 & 0.1290 & 0.6057 & \textcolor{red}{\textbf{0.2143}} & 0.6372 & 0.1818 \\
Pulmonary edema & 0.0076 & 0.8839 & 0.0769 & 0.8385 & \textcolor{red}{\textbf{0.2105}} & 0.8071 & 0.0870 \\
\end{tabular} \hspace{-0.2in}
\caption{AUC and F1 scores for individual diagnostic codes.}
\label{tab:individual-results}
\end{table*}

\begin{table*}[ht]
\centering
\scriptsize
\begin{tabular}{lccccccc}
\multicolumn{8}{c}{\textbf{Classifier Performance on Each Diagnostic Code, Sorted by F1}} \\
\toprule
& & \multicolumn{2}{c}{\textbf{Msmt.}} & \multicolumn{2}{c}{\textbf{Msmt. + indic.}} & \multicolumn{2}{c}{\textbf{Indic.}} \\
\textbf{Condition} & Base rate & AUC & F1 & AUC & \textcolor{red}{\textbf{F1}} & AUC & F1 \\
\midrule
(Acute) pancreatitis & 0.0106 & 0.8712 & 0.0769 & 0.9512 & \textcolor{red}{\textbf{0.2000}} & 0.8182 & 0.0571 \\
Tumor, disseminated or metastatic & 0.0180 & 0.7178 & 0.0938 & 0.7415 & \textcolor{red}{\textbf{0.1967}} & 0.6837 & 0.1062 \\
Hematoma, intracranial & 0.0299 & 0.7724 & \textcolor{red}{\textbf{0.2278}} & 0.8249 & 0.1892 & 0.7518 & 0.1474 \\
Neutropenia (agranulocytosis) & 0.0108 & 0.8285 & 0.0000 & 0.8114 & 0.1852 & 0.8335 & \textcolor{red}{\textbf{0.2609}} \\
Arrhythmia, other & 0.0087 & 0.8536 & 0.0000 & 0.8977 & \textcolor{red}{\textbf{0.1818}} & 0.8654 & 0.0000 \\
Child abuse, suspected & 0.0065 & 0.9544 & \textcolor{red}{\textbf{0.2222}} & 0.8642 & 0.1818 & 0.8227 & 0.0870 \\
Encephalopathy, hypoxic/ischemic & 0.0116 & 0.8242 & 0.1429 & 0.8571 & \textcolor{red}{\textbf{0.1818}} & 0.8009 & 0.0800 \\
Epidural hematoma & 0.0098 & 0.7389 & 0.0455 & 0.8233 & \textcolor{red}{\textbf{0.1818}} & 0.7936 & 0.1000 \\
Tumor, gastrointestinal & 0.0100 & 0.8112 & 0.1429 & 0.8636 & \textcolor{red}{\textbf{0.1778}} & 0.8732 & 0.0984 \\
Craniofacial malformation & 0.0133 & 0.8707 & \textcolor{red}{\textbf{0.2667}} & 0.8514 & 0.1778 & 0.6928 & 0.2286 \\
Gastroesophageal reflux & 0.0182 & 0.7571 & \textcolor{red}{\textbf{0.1818}} & 0.8554 & 0.1690 & 0.7739 & 0.1600 \\
Pneumonia, bacterial (pneumococ.) & 0.0186 & 0.8876 & 0.1333 & 0.8806 & \textcolor{red}{\textbf{0.1600}} & 0.8616 & 0.0000 \\
Pneumonia, undetermined & 0.0179 & 0.8323 & 0.1481 & 0.8269 & \textcolor{red}{\textbf{0.1583}} & 0.7772 & 0.0947 \\
Cerebral edema & 0.0059 & 0.8275 & 0.0000 & 0.9469 & \textcolor{red}{\textbf{0.1538}} & 0.9195 & 0.1500 \\
Pneumonia due to inhalation & 0.0078 & 0.7917 & 0.1111 & 0.8602 & \textcolor{red}{\textbf{0.1538}} & 0.8268 & 0.0566 \\
Metabolic or endocrine disorder & 0.0095 & 0.7718 & 0.0364 & 0.6929 & 0.1538 & 0.6319 & \textcolor{red}{\textbf{0.2000}} \\
Disorder of kidney and ureter, other & 0.0204 & 0.8486 & \textcolor{red}{\textbf{0.2857}} & 0.8650 & 0.1500 & 0.8238 & 0.2500 \\
Urinary tract infection & 0.0137 & 0.7478 & 0.1154 & 0.7402 & \textcolor{red}{\textbf{0.1481}} & 0.7229 & 0.0588 \\
Subdural hematoma & 0.0147 & 0.8270 & \textcolor{red}{\textbf{0.1449}} & 0.8884 & 0.1429 & 0.8190 & 0.0476 \\
Near drowning & 0.0068 & 0.8296 & 0.0741 & 0.7917 & \textcolor{red}{\textbf{0.1404}} & 0.6897 & 0.0397 \\
Cardiac arrest, outside hospital & 0.0118 & 0.8932 & 0.0976 & 0.8791 & \textcolor{red}{\textbf{0.1379}} & 0.8881 & 0.0556 \\
Pleural effusion & 0.0113 & 0.8549 & 0.1081 & 0.8186 & \textcolor{red}{\textbf{0.1351}} & 0.7605 & 0.1151 \\
Bronchopulmonary dysplasia & 0.0252 & 0.8309 & \textcolor{red}{\textbf{0.1915}} & 0.7952 & 0.1304 & 0.8503 & 0.1203 \\
Hyponatremia & 0.0056 & 0.5707 & 0.0187 & 0.7398 & \textcolor{red}{\textbf{0.1176}} & 0.8775 & 0.0000 \\
Suspected septicemia, rule out & 0.0143 & 0.7378 & 0.0923 & 0.7402 & \textcolor{red}{\textbf{0.1029}} & 0.6769 & 0.0000 \\
Thrombocytopenia & 0.0112 & 0.7381 & 0.0822 & 0.7857 & \textcolor{red}{\textbf{0.1026}} & 0.8585 & 0.0800 \\
(Benign) intracranial hypertension & 0.0099 & 0.8494 & 0.0000 & 0.9018 & 0.1020 & 0.8586 & \textcolor{red}{\textbf{0.1224}} \\
Pericardial effusion & 0.0055 & 0.8997 & 0.0870 & 0.9085 & \textcolor{red}{\textbf{0.1017}} & 0.9000 & 0.0714 \\
Pulmonary contusion & 0.0068 & 0.9029 & 0.0606 & 0.8831 & \textcolor{red}{\textbf{0.0984}} & 0.8197 & 0.0225 \\
Surgery, gastrointestinal & 0.0104 & 0.6705 & 0.0714 & 0.6666 & \textcolor{red}{\textbf{0.0976}} & 0.5545 & 0.0233 \\
Respiratory Arrest & 0.0062 & 0.8404 & 0.0000 & 0.8741 & \textcolor{red}{\textbf{0.0952}} & 0.8127 & 0.0444 \\
Trauma, abdominal & 0.0105 & 0.7426 & \textcolor{red}{\textbf{0.1667}} & 0.8623 & 0.0930 & 0.6991 & 0.0426 \\
Atrial septal defect & 0.0107 & 0.7766 & 0.0727 & 0.7765 & \textcolor{red}{\textbf{0.0909}} & 0.7197 & 0.0000 \\
Genetic abnormality & 0.0557 & 0.6629 & \textcolor{red}{\textbf{0.1324}} & 0.6470 & 0.0876 & 0.5705 & 0.1165 \\
Arrhythmia, ventricular & 0.0062 & 0.8532 & 0.0303 & 0.8703 & 0.0870 & 0.8182 & \textcolor{red}{\textbf{0.1250}} \\
Hematologic disorder, other & 0.0114 & 0.6736 & 0.0800 & 0.6898 & \textcolor{red}{\textbf{0.0870}} & 0.8074 & 0.0800 \\
Asthma, stable & 0.0171 & 0.7010 & \textcolor{red}{\textbf{0.0925}} & 0.6607 & 0.0870 & 0.5907 & 0.0741 \\
Neurofibromatosis & 0.0079 & 0.8022 & 0.0469 & 0.7984 & \textcolor{red}{\textbf{0.0816}} & 0.7388 & 0.0160 \\
Tumor, bone & 0.0090 & 0.8830 & 0.0727 & 0.8174 & \textcolor{red}{\textbf{0.0800}} & 0.7649 & 0.0417 \\
Shock, hypovolemic & 0.0088 & 0.7703 & 0.0000 & 0.8433 & \textcolor{red}{\textbf{0.0741}} & 0.8040 & 0.0000 \\
Gastrointestinal bleed, other & 0.0064 & 0.8325 & 0.0541 & 0.7974 & 0.0741 & 0.7996 & \textcolor{red}{\textbf{0.0909}} \\
\end{tabular} \hspace{-0.2in}

\end{table*}

\begin{table*}[ht]
\centering
\scriptsize
\begin{tabular}{lccccccc}
\multicolumn{8}{c}{\textbf{Classifier Performance on Each Diagnostic Code, Sorted by F1}} \\
\toprule
& & \multicolumn{2}{c}{\textbf{Msmt.}} & \multicolumn{2}{c}{\textbf{Msmt. + indic.}} & \multicolumn{2}{c}{\textbf{Indic.}} \\
\textbf{Condition} & Base rate & AUC & F1 & AUC & \textcolor{red}{\textbf{F1}} & AUC & F1 \\
\midrule
Chromosomal abnormality & 0.0173 & 0.8047 & 0.1034 & 0.7197 & 0.0714 & 0.6300 & \textcolor{red}{\textbf{0.1600}} \\
Encephalopathy, other & 0.0093 & 0.8265 & 0.1250 & 0.8736 & 0.0688 & 0.8335 & \textcolor{red}{\textbf{0.1250}} \\
Respiratory syncytial virus & 0.0130 & 0.8876 & \textcolor{red}{\textbf{0.2857}} & 0.9145 & 0.0645 & 0.8716 & 0.0930 \\
(Hereditary) hemolytic anemia, other & 0.0088 & 0.7582 & 0.0548 & 0.8544 & \textcolor{red}{\textbf{0.0645}} & 0.9125 & 0.0513 \\
Obstructive sleep apnea & 0.0185 & 0.7564 & 0.0613 & 0.8200 & 0.0645 & 0.8087 & \textcolor{red}{\textbf{0.1111}} \\
Apnea, central & 0.0142 & 0.7871 & \textcolor{red}{\textbf{0.1600}} & 0.8134 & 0.0625 & 0.8051 & 0.0000 \\
Neuromuscular, other & 0.0132 & 0.7163 & 0.0452 & 0.7069 & \textcolor{red}{\textbf{0.0619}} & 0.6484 & 0.0392 \\
Anemia, acquired & 0.0056 & 0.7378 & \textcolor{red}{\textbf{0.1017}} & 0.7596 & 0.0615 & 0.8129 & 0.0519 \\
Meningitis, bacterial & 0.0070 & 0.4431 & 0.0000 & 0.7676 & \textcolor{red}{\textbf{0.0606}} & 0.5480 & 0.0000 \\
Trauma, long bone injury & 0.0096 & 0.8757 & 0.0952 & 0.9085 & 0.0597 & 0.7946 & \textcolor{red}{\textbf{0.1176}} \\
Bowel (intestinal) obstruction & 0.0104 & 0.7512 & \textcolor{red}{\textbf{0.0984}} & 0.6559 & 0.0597 & 0.6936 & 0.0424 \\
Neurologic disorder, other & 0.0288 & 0.7628 & \textcolor{red}{\textbf{0.1481}} & 0.6978 & 0.0588 & 0.5971 & 0.0769 \\
Panhypopituitarism & 0.0057 & 0.7763 & 0.0000 & 0.7724 & \textcolor{red}{\textbf{0.0571}} & 0.6415 & 0.0000 \\
Thyroid dysfunction & 0.0072 & 0.6310 & 0.0369 & 0.6420 & \textcolor{red}{\textbf{0.0541}} & 0.6661 & 0.0000 \\
Coma & 0.0056 & 0.6483 & \textcolor{red}{\textbf{0.1250}} & 0.6823 & 0.0513 & 0.7155 & 0.0000 \\
Spinal cord lesion & 0.0133 & 0.7298 & \textcolor{red}{\textbf{0.0585}} & 0.7052 & 0.0488 & 0.8168 & 0.0414 \\
Pneumonia, other (mycoplasma) & 0.0188 & 0.8589 & \textcolor{red}{\textbf{0.1613}} & 0.8792 & 0.0476 & 0.8424 & 0.1164 \\
Trauma, blunt & 0.0065 & 0.9156 & \textcolor{red}{\textbf{0.0513}} & 0.8138 & 0.0469 & 0.7426 & 0.0177 \\
Surgery, thoracic & 0.0058 & 0.7405 & 0.0000 & 0.6948 & 0.0469 & 0.6087 & \textcolor{red}{\textbf{0.0909}} \\
Neuroblastoma & 0.0059 & 0.6526 & 0.0306 & 0.7268 & \textcolor{red}{\textbf{0.0360}} & 0.7775 & 0.0346 \\
Obesity & 0.0098 & 0.7503 & 0.0365 & 0.6814 & 0.0351 & 0.6647 & \textcolor{red}{\textbf{0.0667}} \\
Obstructed ventriculoperitoneal shunt & 0.0073 & 0.6824 & 0.0267 & 0.7114 & 0.0331 & 0.7516 & \textcolor{red}{\textbf{0.0667}} \\
Ventricular septal defect & 0.0119 & 0.6641 & \textcolor{red}{\textbf{0.1081}} & 0.5680 & 0.0294 & 0.5593 & 0.0444 \\
Croup Syndrome, UAO & 0.0069 & 0.9418 & 0.2222 & 0.9834 & 0.0000 & 0.9682 & \textcolor{red}{\textbf{0.2222}} \\
Sickle-cell anemia, unspecified & 0.0080 & 0.6262 & 0.0000 & 0.9627 & 0.0000 & 0.8661 & \textcolor{red}{\textbf{0.1250}} \\
Biliary atresia & 0.0063 & 0.9383 & \textcolor{red}{\textbf{0.2667}} & 0.9164 & 0.0000 & 0.7589 & 0.0714 \\
Metabolic acidosis (<7.1) & 0.0083 & 0.9475 & \textcolor{red}{\textbf{0.1818}} & 0.9046 & 0.0000 & 0.9143 & 0.1538 \\
Immunologic disorder, other & 0.0094 & 0.9539 & \textcolor{red}{\textbf{0.1500}} & 0.8868 & 0.0000 & 0.8969 & 0.1212 \\
Pulmonary hypertension, other & 0.0112 & 0.9259 & \textcolor{red}{\textbf{0.2500}} & 0.8826 & 0.0000 & 0.8098 & 0.0000 \\
Trauma, chest & 0.0051 & 0.9261 & 0.0000 & 0.8818 & 0.0000 & 0.7820 & 0.0000 \\
Spinal muscular atrophy & 0.0052 & 0.9666 & 0.0000 & 0.8658 & 0.0000 & 0.8362 & 0.0000 \\
Trauma, unspecified & 0.0065 & 0.7153 & \textcolor{red}{\textbf{0.1481}} & 0.8657 & 0.0000 & 0.8224 & 0.0594 \\
Bone marrow transplant, status post & 0.0097 & 0.8161 & \textcolor{red}{\textbf{0.5217}} & 0.8562 & 0.0000 & 0.8505 & 0.1695 \\
Surgery, orthopaedic  & 0.0180 & 0.7839 & \textcolor{red}{\textbf{0.1029}} & 0.8192 & 0.0000 & 0.7331 & 0.0000 \\
Gastrointestinal bleed, upper & 0.0063 & 0.8388 & 0.0000 & 0.8078 & 0.0000 & 0.7256 & 0.0000 \\
Arrhythmia, supraventricular tachy. & 0.0055 & 0.8178 & \textcolor{red}{\textbf{0.0385}} & 0.7867 & 0.0000 & 0.8199 & 0.0000 \\
Congenital central alveolar hypovent. & 0.0057 & 0.7067 & 0.0000 & 0.7716 & 0.0000 & 0.7282 & 0.0000 \\
Tetralogy of fallot & 0.0061 & 0.5759 & 0.0000 & 0.7614 & 0.0000 & 0.7637 & 0.0000 \\
Cardiac disorder, other & 0.0071 & 0.7229 & \textcolor{red}{\textbf{0.0519}} & 0.7552 & 0.0000 & 0.6287 & 0.0000 \\
Hydrocephalus, shunt failure & 0.0083 & 0.7715 & 0.0000 & 0.7542 & 0.0000 & 0.7986 & \textcolor{red}{\textbf{0.0635}} \\
Cerebral infarction (CVA) & 0.0058 & 0.6766 & 0.0000 & 0.7495 & 0.0000 & 0.7148 & \textcolor{red}{\textbf{0.1333}} \\
Congenital heart disorder, other & 0.0084 & 0.7590 & 0.0000 & 0.7277 & 0.0000 & 0.7803 & \textcolor{red}{\textbf{0.0583}} \\
Gastrointestinal disorder, other & 0.0139 & 0.6755 & 0.0336 & 0.6821 & 0.0000 & 0.6465 & \textcolor{red}{\textbf{0.1026}} \\
Aspiration & 0.0072 & 0.6727 & \textcolor{red}{\textbf{0.0533}} & 0.6734 & 0.0000 & 0.6792 & 0.0333 \\
Dehydration & 0.0105 & 0.7356 & \textcolor{red}{\textbf{0.0690}} & 0.6636 & 0.0000 & 0.5899 & 0.0000 \\
Tumor, thoracic & 0.0077 & 0.6931 & \textcolor{red}{\textbf{0.0513}} & 0.6249 & 0.0000 & 0.6815 & 0.0292 \\
UAO, extubation, status post & 0.0085 & 0.8295 & \textcolor{red}{\textbf{0.0672}} & 0.6063 & 0.0000 & 0.6128 & 0.0000 \\
\bottomrule
\end{tabular} \hspace{-0.2in}

\end{table*}

\clearpage
\section{Missing}
\label{sec:sampling}
In this appendix, we present information 
about the sampling rates and missingness characteristics of our 13 variables. 
The first column lists the average number of measurements per hour in all episodes with at least one measurement (excluding episodes where the variable is missing entirely). The second column lists the fraction of episodes in which the variable is missing completely (there are zero measurements). The third column lists the missing rate in the resulting discretized sequences.\\
\vspace{5pt}

\begin{table}[h!]
\begin{tabular}{lccc}
\toprule
\textbf{Variable} & \textbf{Msmt./hour} & \textbf{Missing entirely} & \textbf{Frac. missing} \\
\midrule
Diabstolic blood pressure  &  0.5162  &  0.0135  &  0.1571\\
Systolic blood pressure  &  0.5158  &  0.0135  &  0.1569\\
Peripheral capillary refall rate  &  1.0419  &  0.0140  &  0.5250\\
End-tidal CO$_2$  &  0.9318  &  0.5710  &  0.5727\\
Fraction inspired O$_2$  &  1.3004  &  0.1545  &  0.7873\\
Total glasgow coma scale  &  1.0394  &  0.0149  &  0.5250\\
Glucose  &  1.4359  &  0.1323  &  0.9265\\
Heart rate  &  0.2477  &  0.0133  &  0.0329\\
pH  &  1.4580  &  0.3053  &  0.9384\\
Respiratory rate  &  0.2523  &  0.0147  &  0.0465\\
Pulse oximetry  &  0.1937  &  0.0022  &  0.0326\\
Temperature  &  1.0210  &  0.0137  &  0.5235\\
Urine output  &  1.1160  &  0.0353  &  0.5980 \\
\bottomrule
\end{tabular}
\caption{Sampling rates and missingness statistics for all $13$ features.}
\end{table}

\end{document}